\documentclass[a4paper]{cas-sc}
\usepackage[numbers,sort&compress]{natbib}
\usepackage{lineno} 
\usepackage{float}
\usepackage{placeins}
\usepackage{xcolor}

\def\tsc#1{\csdef{#1}{\textsc{\lowercase{#1}}\xspace}}
\tsc{WGM}
\tsc{QE}
\tsc{EP}
\tsc{PMS}
\tsc{BEC}
\tsc{DE}
\begin{document}
\let\WriteBookmarks\relax
\def\floatpagepagefraction{1}
\def\textpagefraction{.001}
\shorttitle{}
\shortauthors{Zhu et~al.}

\title [mode = title]{PIS3R: Very Large Parallax Image Stitching via Deep 3D Reconstruction}                      

\affiliation[1]{organization={Hunan University},
                city={Changsha},
                postcode={410082}, 
                state={Hunan},
                country={China}}

\affiliation[2]{organization={Institut Pascal, Universit\'e Clermont Auvergne / CNRS},
                postcode={63178}, 
                city={Clermont-Ferrand},
                country={France}}

\author[1]{Muhua Zhu}[orcid=0009-0005-9049-2321]
\fnmark[1]
\ead{casmyzhu@hnu.edu.cn}

\credit{Writing – original draft, Methodology, Data curation, Software, Validation}

\author[1]{Xinhao Jin}[orcid=0009-0006-5722-2896]
\fnmark[1]
\ead{jinxinhao@hnu.edu.cn}
\credit{Data curation, Validation, Visualization, Formal analysis}

\author[1]{Chengbo Wang}[orcid=0009-0009-1385-1633]
\ead{wangchb@hnu.edu.cn}
\credit{Writing - Original draft preparation, Validation, Formal analysis}

\author[2]{Yongcong Zhang}[orcid=0009-0009-1254-6549]
\ead{yongcongzhang@hnu.edu.cn}
\credit{Validation ,Formal analysis}

\author[1]{Yifei Xue}[orcid=0000-0002-4443-4367]
\ead{iflyhsueh@gmail.com}
\credit{Supervision}

\author[1]{Tie Ji}[orcid=0009-0005-0822-0600]
\ead{jietie\_hnu@163.com}
\credit{Supervision, Funding acquisition}
\cormark[1]

\author[1]{Yizhen Lao}[orcid=0000-0002-6284-1724]
\ead{yizhenlao@hnu.edu.cn}
\credit{Conceptualization, Project administration, Writing – review & editing}
\cormark[2]

\cortext[cor1]{Corresponding author}
\cortext[cor2]{Principal corresponding author}
\fntext[fn1]{Two authors contributed equally to this work.}

\begin{abstract}
Image stitching aims to align two images taken from different viewpoints into one seamless, wider image. However, when the 3D scene contains depth variations and the camera baseline is significant, noticeable parallax occurs—meaning the relative positions of scene elements differ substantially between views. Most existing stitching methods struggle to handle images with such large parallax effectively. To address this challenge, in this paper, we propose an image stitching solution called \textbf{PIS3R} that is robust to very large parallax(VLP) based on the novel concept of deep 3D reconstruction. First, we apply visual geometry grounded transformer to two input images with VLP to obtain both intrinsic and extrinsic parameters, as well as the dense 3D scene reconstruction. Subsequently, we reproject the reconstructed dense point cloud onto a designated reference view using the recovered camera parameters, achieving pixel-wise alignment and generating an initial stitched image. Finally, to further address potential artifacts such as reprojected bias or noise in the initial stitching, we propose an image diffusion module to obtain the refined result.
Compared with existing methods, our solution is VLP tolerant and also provides results that largely preserve the geometric integrity of all pixels in the 3D photogrammetric context, enabling direct applicability to downstream 3D vision tasks such as SfM. Experimental results demonstrate that the proposed algorithm provides accurate stitching results for images with VLP, and outperforms existing methods qualitatively and quantitatively.
\end{abstract}


\maketitle
\section{Introduction}
Image stitching is a useful technique used to generate a wide field-of-view (FoV) scene by combining multiple images captured with limited FoV. It plays a vital role across various domains, including autonomous driving \cite{auto_driveing_kinzig2022real}, medical imaging, surveillance, and virtual reality \cite{VR_anderson2016jump}.

Most image stitching approaches follow a standard pipeline that begins by detecting and matching feature points among the input image pair. Based on these matches, global parametric transformations—typically homographies—are estimated to warp one image onto the coordinate system of the other. The final stitched result is then obtained by blending the warped image with the reference image, particularly resolving pixel values in overlapping regions.

\begin{figure}
    \centering
    \includegraphics[width=\textwidth]{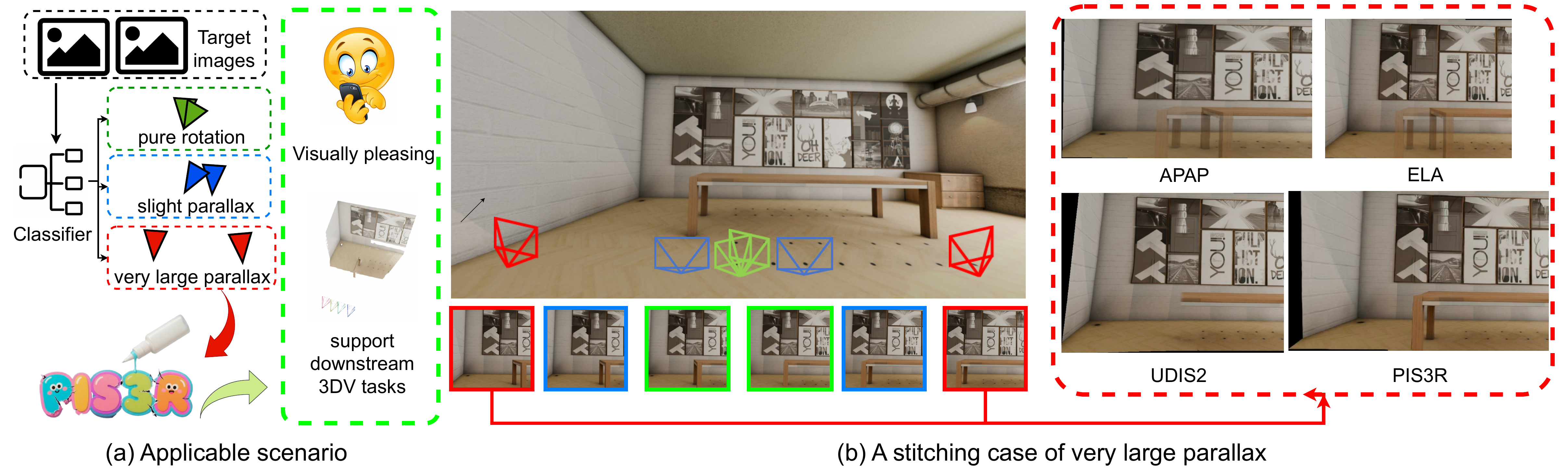}
    \caption{
    PIS3R for very large parallax image stitching.
    (a) The image stitching task can be classified (Sec. C in Appendix) into three cases based on the baseline between target images: \textit{pure rotation}, \textit{slight parallax}, and \textit{very large parallax}. In this paper, we propose a novel image stitching solution named PIS3R, specifically for stitching images with \textit{very large parallax}. (b) Representative existing image stitching methods~\cite{apap_zaragoza2014projective,ELA_li2018parallax,udis2_nie2023parallax} are unable to handle stitching tasks involving \textit{very large parallax}. Instead, our solution is achieving promising results in such challenging scenario.
    }
    \label{fig:top_picture}
\end{figure}

Image warping is a key challenge in image stitching. Homography, a widely used planar transformation model, assumes a flat scene \cite{hartley2003multiple}. However, in real-world scenes with depth variation and large camera baselines, this assumption fails, leading to parallax effects and noticeable misalignments—especially near object boundaries as shown in Fig.~\ref{fig:top_picture}.
\subsection{Motivations}
\noindent \textbf{Very large parallax (VLP).} To mitigate parallax artifacts in image stitching, 
adaptive warping methods have been developed that divide images into grids or pixels and apply locally varying transformations \mbox{\cite{ELA_li2018parallax,lin2011smoothly,apap_zaragoza2014projective}}. 
Energy minimization frameworks further optimize these warps to reduce geometric distortion \cite{ELA_li2018parallax, lin2011smoothly, zhang2016multi}. 
Some approaches employ seam-cutting techniques to align local regions and conceal misalignments in others \cite{gao2013seam, lin2016seagull, PTIS_Zhang_2014_CVPR}. However, existing smooth warping methods~\mbox{\cite{ELA_li2018parallax,lin2011smoothly,apap_zaragoza2014projective}}
often fail under VLP, where adjacent pixels in one image may not correspond to nearby pixels in the other view. While video-based methods using epipolar geometry have been proposed \cite{lee2018video}, they rely on temporal motion cues and are not directly applicable to still image stitching. 

\noindent \textbf{Downstream 3D vision (3DV) tasks.}  Almost all existing image stitching methods, including adaptive warping, seam cutting, and deep image stitching, aim to seamlessly merge two input images, with visual quality as the primary criterion. However, they generally overlook whether the stitched result preserves the underlying 3D projection geometry while this geometric consistency is critical for enabling downstream 3D vision tasks such as structure-from-motion, SLAM, visual pose estimation, and depth estimation.

\textbf{In summary,} if we can stitch images with VLP and produce not only visually seamless result but also are capable to support downstream 3DV tasks. 

\subsection{Contribution}

As shown in Fig.~\ref{fig:parallax_exp} and Table~\ref{tab:comparison}, to address the limitations of existing methods under extreme parallax, we propose PIS3R, a novel image stitching framework leveraging deep 3D reconstruction for robust alignment in very large parallax (VLP) scenarios. Our method first applies the feed-forward scene reconstruction model Visual Geometry Grounded Transformer (VGGT)~\cite{wang2025vggt} to uncalibrated image pairs, jointly recovering camera intrinsics, extrinsics, and a dense 3D scene. The reconstructed point cloud is then reprojected onto a reference view using the estimated poses, producing an initial stitched image with pixel-wise geometric alignment. A diffusion module further refines this result, filling holes and suppressing reprojection artifacts.

Unlike prior work that prioritizes visual seamlessness, PIS3R explicitly preserves 3D photogrammetric consistency for every pixel, enabling high-quality stitching under large parallax and ensuring compatibility with downstream 3D vision tasks like SfM, SLAM, and depth estimation. Experiments show PIS3R outperforms existing methods both qualitatively and quantitatively. Our contributions are:

\begin{itemize}
  \item We introduce a comprehensive image stitching dataset specifically designed for very large parallax (VLP) scenarios, comprising both high-quality synthetic scenes and real-world captures, to address the lack of VLP data in existing benchmarks.
  \item We propose a novel pipeline for stitching images with VLP based on the deep 3D reconstruction results and followed by a diffusion-based refinement, called PIS3R, which is a fundamental departure from the technical pipeline of existing methods. 
  \item The idea of applying deep 3D reconstruction for image stitching that allows the produced results to achieve not only visually seamless alignment but also maintain 3D projection consistency, which enables our method to better support downstream 3D tasks such as SfM and SLAM over SOTAs. 
\end{itemize}

\begin{table*}[h]
    \centering
    \caption{Comparison of the proposed method PIS3R, vs. the state-of-the-art parallax-tolerate solutions. Although existing methods can generate stitched images when processing VLP cases, they suffer from significant distortions and artifacts (Fig.~\ref{fig:top_picture}). As a result, they fail to achieve visual pleasantness and maintain structural integrity. Therefore, we argue that current approaches cannot effectively handle VLP scenarios.}
    \label{tab:comparison}

    \resizebox{0.92\textwidth}{!}{%
    \begin{tabular}{cccccccccc}
        \toprule
        \multirow{3}{*}{Method} & \multicolumn{4}{c}{Adaptive Warping} & \multicolumn{2}{c}{Seam Driven} & \multicolumn{2}{c}{Deep Learning}
        \\
        \cmidrule(lr){2-5} \cmidrule(lr){6-7} \cmidrule(lr){8-9}
         & APAP\cite{apap_zaragoza2014projective}
         & DFW\cite{li2015dualfeature}
         & ELA\cite{ELA_li2018parallax}
         & OBJ-GSP\cite{obj-gsp_cai2025object}
         & PTIS\cite{PTIS_Zhang_2014_CVPR}
         & SEAGULL\cite{lin2016seagull}
         & UDIS\cite{udis_d_nie2021unsupervised}
         & UDIS++\cite{udis2_nie2023parallax}
         & PIS3R\\

        & \textcolor{blue}{TPAMI'14}
        & \textcolor{blue}{ICCV'15}
        & \textcolor{blue}{TMM'18}
        & \textcolor{blue}{AAAI'25}
        & \textcolor{blue}{CVPR'14}
        & \textcolor{blue}{ECCV'16}
        & \textcolor{blue}{TIP'21}
        & \textcolor{blue}{ICCV'23}
        \\
        \midrule
        Pure rotation
        & \checkmark & \checkmark & \checkmark & \checkmark
        & \checkmark & \checkmark & \checkmark & \checkmark & \checkmark\\

        Slight parallax
        & \checkmark & \checkmark & \checkmark & \checkmark
        & \checkmark & \checkmark & & \checkmark & \checkmark\\

        Very large parallax
        & & & & & & & & & \checkmark\\
        \bottomrule
    \end{tabular}%
    }

\end{table*}

\section{Related Work}
\noindent \textbf{Traditional image stitching.} \textbf{\textit{1)}} \textbf{\textit{Adaptive warping.}} Traditional stitching methods often leverage discriminative features like SIFT \cite{lowe2004distinctive} for estimating flexible warps such as DHW \cite{gao2011dual}, APAP \cite{apap_zaragoza2014projective}, and TFA \cite{lin2011smoothly}. Extensions like DFW \cite{li2015dualfeature}, GES-GSP \cite{du2022gesgsp} and OBJ-GSP \cite{obj-gsp_cai2025object} incorporate lines, edges, mesh or semantic planes to enhance structural preservation. Furthermore, semantic planar priors can improve local alignment \cite{li2021semanticplanar}, and multiple registrations can accommodate regions at different depths \cite{herrmann2018multiple}.
\textbf{\textit{2)}} \textbf{\textit{Seam cutting.}} Seam-cutting is commonly employed as a post-processing step for artifact reduction. Methods define energy terms based on photometric, gradient, or motion cues \cite{kwatra2003graphcut, agarwala2004interactive, bao2013fast}, and minimize them via graph cuts to produce visually coherent results. Seam-based alignment optimization and object-aware seam selection have also been explored to reduce stitching artifacts \cite{lin2016seagull, PTIS_Zhang_2014_CVPR, herrmann2018object}. \textbf{However}, these geometry-driven methods face two limitations: 1) they are sensitive to missing or weak structural features, leading to alignment failure; and 2) when geometric features are abundant, the computational overhead increases significantly.

\begin{figure}
    \centering
    \includegraphics[width=0.7\columnwidth]{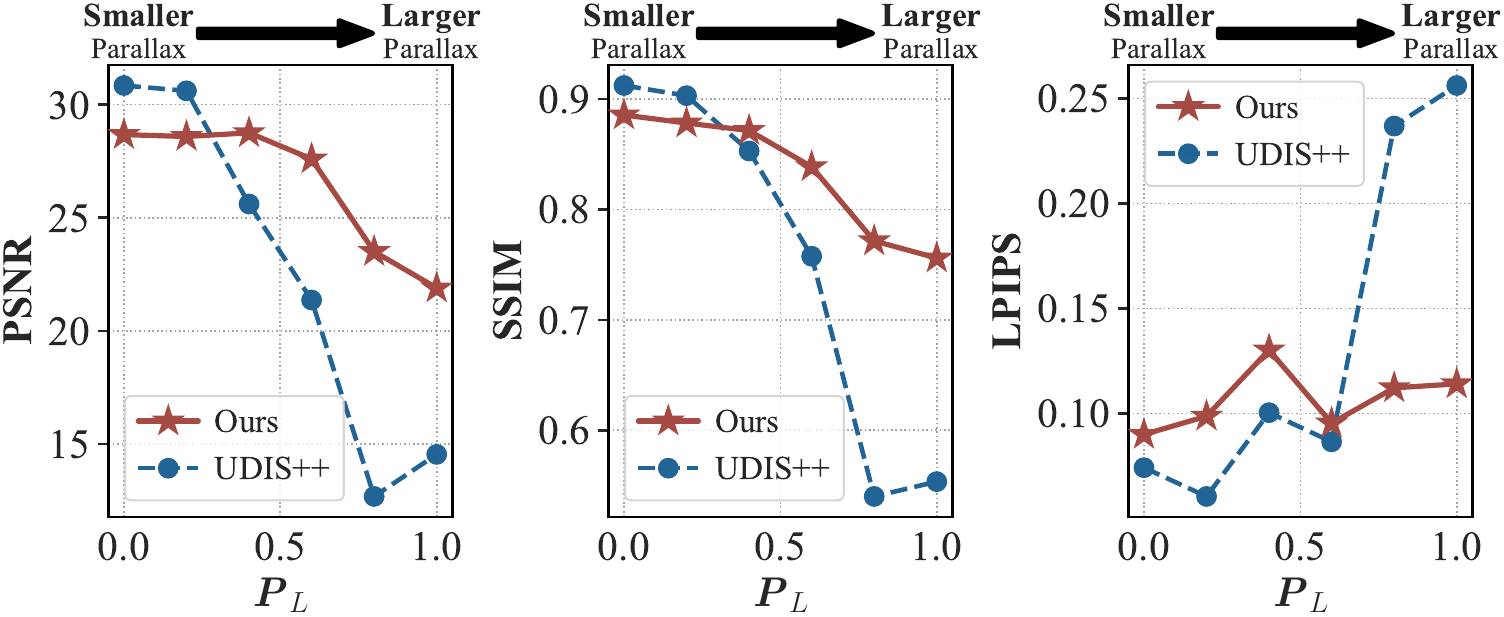}
    \caption{
    The relationship between quality of stitched image and parallax variation.
    Our PIS3R method maintains significantly higher stability than UDIS++ across all three metrics (PSNR, SSIM, and LPIPS). 
    As Parallax level $P_L$ (Sec. C, Appendix) increases, stitched image quality from UDIS++ degrades substantially, 
    whereas PIS3R demonstrates superior robustness to parallax variations. Moreover, PIS3R delivers comparable visual quality to state-of-the-art methods under pure rotation and slight parallax.
    }
    \label{fig:parallax_exp}
\end{figure}

\noindent \textbf{Deep image stitching.} These approaches learn high-level semantic representations from data. They have been developed under supervised
\mbox{\cite{nie2022edgepreserved,song2021multihomography,kweon2023pixelwise}}, weakly supervised \cite{song2022weakly}, and unsupervised paradigms \cite{jin2025pixelstitch}, demonstrating strong adaptability to complex scenes. Among these, unsupervised methods are particularly appealing due to the scarcity of ground truth stitched images. \textbf{However}, most of them rely on homography-based alignment, which struggles under very large parallax as shown in Fig.~\ref{fig:parallax_exp}, often resulting in visible blur and artifacts in the reconstructed regions without mentioning being unable to preserve the 3D projection consistency. Recent low-light image enhancement methods improve image visibility and perceptual quality under challenging illumination conditions~\cite{wang2026eicenet,shakibania2025cdan}. These methods are complementary to PIS3R, which focuses on depth-dependent geometric misalignment rather than photometric degradation.

\noindent
\textbf{3D geometry and depth-aware image stitching.}
Several methods exploit epipolar geometry, disparity, or depth priors to handle parallax.
\cite{lee2018video}
introduces adaptive pixel warping based on epipolar geometry for large-parallax stitching, while
\cite{lin2022image}
uses disparity to perform multi-plane alignment.
\cite{liao2025natural}
further combines depth maps with epipolar geometry to construct piece-wise local homographies, and
\cite{yu2023parallax}
constrains pixel warping using an epipolar displacement field. However, these methods mainly employ 3D cues to guide or constrain 2D warping, rather than explicitly reconstructing and reprojecting the 3D scene, limiting their robustness under very large parallax and severe occlusions.

\section{Methodology}
\label{sec:methodology}

\begin{figure*}
    \centering
    \includegraphics[width=\textwidth]{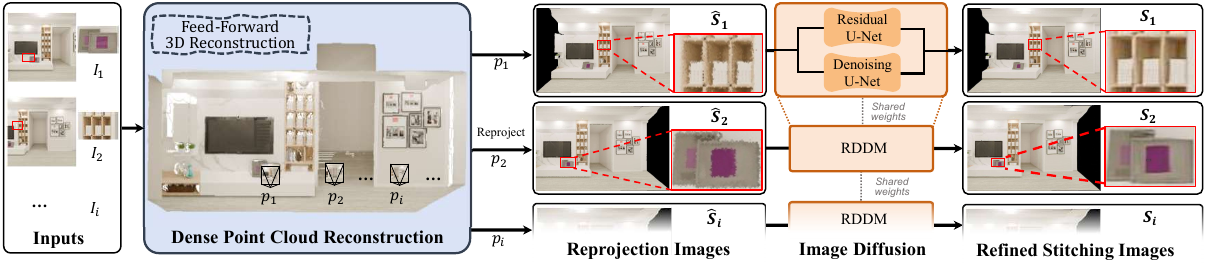}
\caption{
    Overview of the PIS3R pipeline.
    Given a sparse set of input images, we first construct a dense point cloud representation $P_{w}$ using a feed-forward deep 3D reconstruction model.
    Subsequently, the point cloud is reprojected onto the input camera poses~$p_i$ to generate preliminary stitching images~$\hat{S}_i$. 
    While these images preserve the majority of the scene's structural information, they concurrently introduce substantial noise artifacts. 
    To address this limitation, we train a denoising diffusion model to restore image fidelity, ultimately producing refined stitching images~$S_i$.
    }  
    \label{fig:pipeline}
\end{figure*}
In this section, we begin with a preliminary analysis of parallax induced by camera motion, followed by an overview of our image stitching framework, PIS3R. We then describe each component in detail and finally present the implementation details.

\subsection{Preliminary: Parallax Produced by Camera Motion}

Following the standard two-view projective geometry~\cite{hartley2003multiple}, consider a 3D point $\mathbf{P}$ observed in two views with relative camera motion $(R,\mathbf{t})$. Its projections satisfy:
\begin{equation}
s_1\mathbf{p}_1 = K\mathbf{P}, \qquad
s_2\mathbf{p}_2 = K(R\mathbf{P}+\mathbf{t}),
\label{eq:two_view_projection}
\end{equation}
where $\mathbf{p}_1$ and $\mathbf{p}_2$ are the homogeneous pixel coordinates, $K$ is the camera intrinsic matrix, $R$ and $\mathbf{t}$ are the relative rotation and translation, and $s_1$ and $s_2$ denote the corresponding depths. From $\mathbf{P}=s_1K^{-1}\mathbf{p}_1$, we obtain:
\begin{equation}
\mathbf{p}_2
\sim
K\left(
s_1RK^{-1}\mathbf{p}_1+\mathbf{t}
\right),
\label{eq:parallax_general}
\end{equation}
which describes the joint effects of camera motion and scene depth on inter-view correspondence. For pure rotation, Eq.~\eqref{eq:parallax_general} reduces to
$\mathbf{p}_2 \sim KRK^{-1}\mathbf{p}_1.$
The depth term is eliminated, indicating that rotation-induced pixel displacement is depth-independent and can be represented by a global homography.

In contrast, translation introduces explicit depth dependence. Let $\mathbf{q}_1=K^{-1}\mathbf{p}_1=(x,y,1)^\top$, $\mathbf{t}=(t_x,t_y,t_z)^\top$, and denote the three rows of $R$ by $\mathbf{r}_1^\top$, $\mathbf{r}_2^\top$, and $\mathbf{r}_3^\top$. The normalized coordinates in the second view are
\begin{equation}
x_2 =
\frac{
s_1\mathbf{r}_1^\top\mathbf{q}_1+t_x
}{
s_1\mathbf{r}_3^\top\mathbf{q}_1+t_z
},
\qquad
y_2 =
\frac{
s_1\mathbf{r}_2^\top\mathbf{q}_1+t_y
}{
s_1\mathbf{r}_3^\top\mathbf{q}_1+t_z
}.
\label{eq:normalized_projection}
\end{equation}
Their dependence on scene depth can be measured by
\begin{equation}
\frac{\partial x_2}{\partial s_1}
=
\frac{
(\mathbf{r}_1^\top\mathbf{q}_1)t_z
-
t_x(\mathbf{r}_3^\top\mathbf{q}_1)
}{
\left(
s_1\mathbf{r}_3^\top\mathbf{q}_1+t_z
\right)^2
},
\qquad
\frac{\partial y_2}{\partial s_1}
=
\frac{
(\mathbf{r}_2^\top\mathbf{q}_1)t_z
-
t_y(\mathbf{r}_3^\top\mathbf{q}_1)
}{
\left(
s_1\mathbf{r}_3^\top\mathbf{q}_1+t_z
\right)^2
}.
\label{eq:depth_derivative_xy}
\end{equation}
Both derivatives are zero under pure rotation, but are generally non-zero when translation exists. Thus, translation is the source of depth-dependent parallax, while rotation affects its projected direction and magnitude.

For the intuitive case of pure horizontal translation, $R=I$ and $\mathbf{t}=(B,0,0)^\top$, the horizontal pixel displacement becomes
\begin{equation}
|\Delta u|
=
\frac{f_x|B|}{s_1},
\label{eq:parallax_baseline}
\end{equation}
where $\Delta u$ is the horizontal pixel displacement, $B$ is the camera baseline, and $f_x$ is the horizontal focal length. Thus, the displacement increases with the baseline and decreases with scene depth. For foreground and background points with depths $s_f$ and $s_b$, respectively, their displacement difference is
\begin{equation}
|\Delta u_f-\Delta u_b|
=
f_x|B|
\left|
\frac{1}{s_f}
-
\frac{1}{s_b}
\right|,
\label{eq:parallax_depth_difference}
\end{equation}
where $\Delta u_f$ and $\Delta u_b$ are their horizontal displacements. This relation shows that parallax increases with the camera baseline and the inverse-depth difference, especially around foreground-background boundaries. Such depth-dependent motion is difficult to represent accurately using a common 2D transformation, motivating the explicit 3D reconstruction and reprojection in PIS3R.

\subsection{PIS3R Pipeline}
\label{sec:PIS3R_Pipeline}
Fig.~\ref{fig:pipeline} illustrates the overall architecture of our PIS3R framework. PIS3R is composed of three components as follows:

\textbf{Step 1:} 3D reconstruction. Given a sparse set of two or more images, we first establish their point cloud representation using a dense stereo model, as illustrated in Sec.~\ref{sec:Deep-based 3D Reconstruction}. 

\textbf{Step 2:} Point Cloud Reprojection. Subsequently, we reproject the point cloud onto the input camera poses predicted by the dense stereo model to obtain a preliminary stitched image, as illustrated in Sec.~\ref{sec:Reprojection}. 

\textbf{Step 3:} Refinement of Reprojected Images. To achieve visually pleasing results while preserving the geometric structure of the scene, we train an image diffusion model to refine the noisy preliminary stitched image, as depicted in Sec.~\ref{sec:RDDM}.

\subsection{Deep-based 3D Reconstruction}
\label{sec:Deep-based 3D Reconstruction}
As shown in Fig.~\ref{fig:pipeline}, we first construct a point cloud representation from the input images to obtain information about the entire scene. Certainly different from prior approaches that achieve image stitching through geometry-based 3D reconstruction, we adopts a deep-based reconstruction framework. Specifically, we employ \textbf{VGGT} \cite{wang2025vggt}\footnote{https://github.com/facebookresearch/vggt} to reconstruct the entire scene and estimate camera parameters simultaneously. The input consists of a sequence of RGB images capturing the same 3D scene $\left(I_{i}\right)_{i=1}^{N}{\in \mathbb{R}^{{H} \times W \times 3}}$ to corresponding 3D annotation information via a transformer, which can be described as:
\begin{equation}
    F\left(\left(I_{i}\right)_{i=1}^{N}\right) = \left(\mathbf{g}_{i}, D_{i}, P_{i}, E_{i}\right)_{i=1}^{N},
    \label{eq:one}
\end{equation}
where $\mathbf{g}_{i} \in \mathbb{R}^9$ is the camera intrinsic and extrinsic parameters, its depth map $D_{i} \in \mathbb{R}^{H \times W}$ and point map $P_{i} \in\mathbb{R}^{3 \times H \times W}$ and a grid $E_{i} \in\mathbb{R}^{C \times H \times W}$.
\begin{figure}
    \centering
    \includegraphics[width=0.7\columnwidth]
    {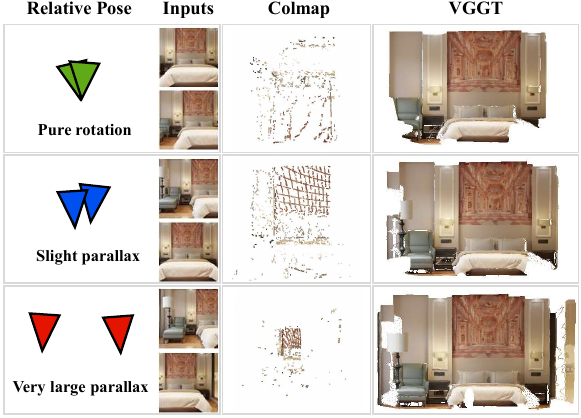}
    \caption{Visual Comparison of 3D Reconstruction Results between COLMAP and VGGT. Reconstruction is performed under three distinct camera pose scenarios: pure rotation, slight parallax, and very large parallax. It can be observed that COLMAP produces significantly sparser point clouds across all scenarios, capturing limited structural information. In contrast, VGGT generates dense point clouds that preserve the majority of scene details.}
    \label{fig:vggt_colmap_result}
\end{figure}
As shown in Fig.~\ref{fig:vggt_colmap_result}, we compare the point cloud representations generated by the geometry-based reconstruction framework COLMAP ~\cite{colmap_schoenberger2016mvs,colmap_schoenberger2016sfm}\footnote{https://github.com/colmap/colmap} and VGGT under three different input settings: pure rotation, slight parallax, and very large parallax. The reconstruction performance of COLMAP under sparse viewpoints is predictable, while the reconstruction results of VGGT are remarkable. The quality of the point cloud and the accuracy of the predicted camera poses are crucial for image stitching using 3D reconstruction. This directly determines the geometric structure and quality of the final stitched image.

\subsection{Image Stitching via Point Cloud Reprojection}
\label{sec:Reprojection}

As shown in Fig.~\ref{fig:pipeline}, to obtain the reprojected image $\hat{S}_i$ from all input images, we first process the global point cloud $P_{w}$. To mitigate ``flying points'' artifacts commonly generated by deep 3D reconstruction models, we apply a depth-adaptive K-Nearest Neighbors filter to prune geometric outliers. Subsequently, the refined point cloud is converted from the world coordinate system to the camera coordinate system $O_c^i$ using the camera extrinsic parameters to generate the transformed point cloud $P_{c}^i \in \mathbb{R}^{N \times 3}$~\cite{hartley2003multiple}. The transformation can be described as:
\begin{equation}
    P_{c}^i = R^i P_{w} + t^i, 
    \label{eq:two}
\end{equation}
where $R^i \in \mathbb{R}^{3 \times 3}$ is the rotation matrix and $t^i \in \mathbb{R}^{3}$ is the translation vector for camera $i$. To ensure geometric validity during projection, we discard invalid points that fall behind the camera plane where $Z_c^i \le 0$.

Subsequently, we dynamically determine the spatial resolution defined by $H_i$ and $W_i$ for the projected canvas to avoid excessive memory consumption while capturing all valid geometric regions. We reproject the valid 3D points in $P_c^i$ onto the 2D image plane to compute a bounding box bounded by $[x_{\min}, x_{\max}]$ and $[y_{\min}, y_{\max}]$ using the original intrinsic matrix $K_i$. To align the point cloud precisely with the cropped canvas, we dynamically adjust the camera intrinsic matrix by shifting its principal point:
\begin{equation}
    \tilde{K}_i = K_i - \begin{bmatrix} 0 & 0 & x_{\min} \\ 0 & 0 & y_{\min} \\ 0 & 0 & 0 \end{bmatrix}.
\end{equation}
This shift ensures the projected points are seamlessly mapped to a valid grid of size $H_i = y_{\max} - y_{\min} + 1$ and $W_i = x_{\max} - x_{\min} + 1$.

To generate a dense reprojected image from the sparse discrete points while avoiding background bleeding artifacts, we utilize a point-based renderer PyTorch3D~\cite{ravi2020accelerating}. Each 3D point is assigned a finite support radius and rasterized onto the dynamic image plane using the adjusted intrinsic matrix $\tilde{K}_i$. For each pixel location $(x,y)$, the rasterizer strictly identifies the closest intersecting point along the $Z$-axis. The pixel is then directly assigned the color of this nearest point, completely bypassing soft blending computations. This deterministic process outputs both the rendered foreground color $C_i \in \mathbb{R}^{H_i \times W_i \times 3}$ and a discrete binary valid mask $M_{val}^i \in \{0, 1\}^{H_i \times W_i}$, which is only used for filling invalid pixels before refinement.

Finally, the preliminary reprojected image $\hat{S}_i \in \mathbb{R}^{H_i \times W_i \times 3}$ is obtained by filling the unprojected void regions with a predefined background color $\mathbf{b}$ (e.g., black) guided by the binary mask:
\begin{equation}
    \hat{S}_i(x, y) = 
    \begin{cases} 
        C_i(x, y), & \text{if } M_{val}^i(x, y) = 1 \\ 
        \mathbf{b}, & \text{otherwise} 
    \end{cases}
\end{equation}

By employing this hard splatting rendering strategy, we naturally obtain a robust geometrically pose-aligned image, providing coarse 3D guidance for stitching. However, sparse inputs still yield incomplete point clouds with structural holes and occlusions, leading to missing regions and misalignments in the reprojected image. To address this, we employ an image diffusion model to produce a visually coherent stitched result.

\subsection{Diffusion-based Stitching Refinement}
\label{sec:RDDM}
\begin{figure*}
    \centering
    \includegraphics[width=0.8\textwidth]{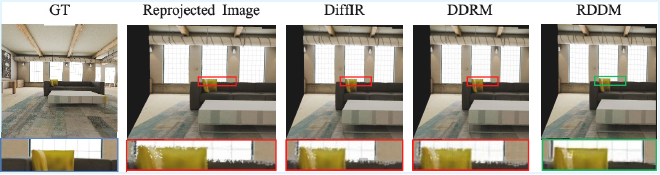}
    \caption{Qualitative comparison of different image restoration diffusion models. We evaluated the performance of several state-of-the-art image restoration diffusion models in refining reprojected images, including \textit{DiffIR}~\cite{diffir_xia2023diffir}, \textit{DDRM}~\cite{ddrm_kawar2022denoising}, and \textit{RDDM}~\cite{liu2024residual}.}
    \label{fig:ablation_diffusion}
\end{figure*}

As illustrated in Fig.~\ref{fig:pipeline}, to improve the quality of the reprojected image, our goal is to learn a conditional distribution $x\sim p(x | \hat{S})$ that generates a high-quality image from the degraded reprojected image. As shown in Fig.~\ref{fig:ablation_diffusion} and Sec.~\ref{sec:ablation_studies}, we compared three image restoration diffusion models. Specifically, we employ \textbf{Residual Denoising Diffusion Models (RDDM)} ~\cite{liu2024residual} modifies to refine reprojected image~$\hat{S}$. Different from traditional diffusion model, RDDM extends $x_T = \varepsilon$ to $x_T = x_{\text{in}} + \varepsilon$ and changes the forward diffusion process by incorporating a mixture of three terms (i.e., $x_0$, $x_{\text{res}}$, and $\varepsilon$):
\begin{flalign}
    x_t &= x_{t-1} + \alpha_t x_{\text{res}} + \beta_t \varepsilon_{t-1},
    \label{eq:five}  \\
    x_t &= x_{0} + \bar\alpha_t x_{\text{res}} + \bar\beta_t \varepsilon_{t},
    \label{eq:six}
\end{flalign}
where $\alpha_t$ and $\beta_t$ are coefficients that respectively control the residual and noise injection, $\bar{\alpha}_t=\sum_{i=1}^t \alpha_i$, $\bar{\beta}_t=\sqrt{\sum_{i=1}^t \beta_i^2}$, $ x_{\text{res}} = x_{\text{in}} - x_0 $, and $\varepsilon \sim \mathcal{N}(0, I)$. 
The reverse process can be formulated as $p_{\theta}(x_{t-1} | x_t) := q_{\sigma}(x_{t-1} | x_t, x_0^{\theta}, x_{\text{res}}^{\theta})$, where $x^{\theta}_{0}$, $x^{\theta}_{res}$ are respectively predicted by a residual U-Net and a denoising U-Net. And estimated target image $x^{\theta}_0=x_t - \bar\alpha_t x^{\theta}_{res}-\bar\beta_t \varepsilon_\theta$ can be recovered according to Eq.~\ref{eq:six}. Subsequently, residual U-Net and denoising U-Net are optimized by the diffusion loss: 
\begin{flalign}
    \min _{res(\theta)}&= \mathbb{E} \left[ \lambda_{\text{res}} \left\| x_{\text{res}} - x_{\text{res}}^{\theta}(x_t, t, x_{\text{in}}) \right\|^2 \right], 
    \label{eq:ten} \\
    \min _{\varepsilon(\theta)} &= \mathbb{E} \left[ \lambda_{\epsilon} \left\| \epsilon - \epsilon_{\theta}(x_t, t, x_{\text{in}}) \right\|^2 \right]
    \label{eq:eleven}
\end{flalign}

RDDM begins with residual information that captures the differences between the target and degraded images. Thus, it can effectively guide the restoration process, maximally preserve the original 3D information provided by the reprojected image, and better maintain the consistency of the geometric structure.

\subsection{Implementation}
\label{sec:Implementation}
\noindent $\bullet$ \textbf{3D reconstruction.} We directly use the open-source pretrained model of VGGT to complete 3D reconstruction. To ensure the integrity of the scene information, we accept all predicted points without applying a confidence threshold.\\
\noindent $\bullet$ \textbf{Diffusion model.}
To train the RDDM model, we created paired training data consisting of degraded and corresponding ground-truth images. Specifically, for each ground-truth image from the \textbf{RainDrop} dataset~\cite{raindrop_qian2018attentive}, we reconstructed its depth map and RGB-D point cloud. The point cloud was then reprojected to the original camera viewpoint, thereby introducing reconstruction and reprojection errors into the rendered image. Before rendering, we smoothed the reconstructed depth map to simulate persistent flying-point artifacts. The resulting reprojected image was used as the degraded input, while the original image served as the supervision target. These paired images constitute the customized 3DIS-D dataset.

We adopted a progressive training strategy with a global learning rate of $8\times10^{-5}$ and batch size of $1$. In the first stage, we trained on 3DIS-D at $512 \times 512$ resolution with $10$ sampling steps for $120,000$ iterations, using ground truth as targets and processed images as degraded inputs. In the second stage, we fine-tuned on the \textbf{GoPro}~\cite{gopro_nah2017deep} dataset at the same resolution with $5$ sampling steps for $20,000$ iterations. Notably, our approach achieves satisfactory results with limited training data. All training and testing were performed on an NVIDIA RTX $4090$ GPU.

\section{Discussion}
\label{sec:Discussion}
\noindent $\bullet$ \textbf{\textit{Is PIS3R suitable for pure rotation or slight parallax?}}
PIS3R delivers comparable visual quality to state-of-the-art methods under pure rotation or slight parallax (Fig. \ref{fig:UDIS-D_dataset_result}) but requires more processing time per image than UDIS++ (Table~\ref{tab:stitching_time}). We recommend PIS3R for scenarios with large parallax and propose a parallax classifier to estimate input image parallax levels.

\noindent $\bullet$ \textbf{\textit{Why deep 3DR?}} Deep 3DR solutions such as VGGT demonstrate significantly greater robustness and accuracy than classical SfM like COLMAP in estimating camera intrinsic and extrinsic parameters, particularly in challenging two-view scenarios such as pure rotation and VLP (as shown in Fig.~\ref{fig:top_picture} and detailed in the Sup material). Moreover, the point maps predicted by VGGT offer substantially higher density compared to traditional SfM point clouds, providing strong support for the following image stitching reprojection step.

\noindent $\bullet$ \textbf{\textit{Why PIS3R is geometry-preserve?}}
PIS3R employs a deep-based reconstruction framework to accurately predict the point cloud representation and the camera intrinsic and extrinsic parameters. Accurate pose estimation of the images is fundamental to preserving geometric structure in PIS3R. This is also the reason why SOTA methods fail to maintain geometric structure.As illustrated in Sec.~\ref{sec:geometry_consistence} and Table~\ref{tab:epipolar_error}.

\begin{figure}
    \centering
    \includegraphics[width=0.5\columnwidth]{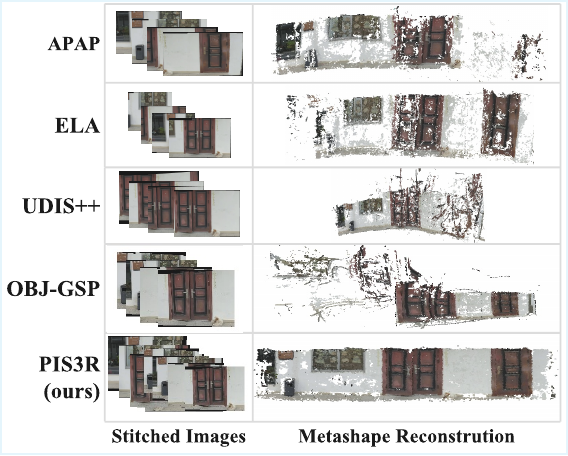}
    \caption{The results of MetaShape reconstruction using stitched images. When used as inputs for 3D reconstruction in MetaShape, images stitched with APAP and ELA resulted in a large number of surface holes. 
    Images stitched with UDIS++ and OBJ-GSP exhibited significant distortions to geometric structure. 
    In contrast, reconstructions generated from our stitched images consistently exhibited high geometric consistency and completeness within MetaShape. 
    This enhanced structural integrity suggests that PIS3R is more suitable for downstream 3D reconstruction tasks.
    }
    \label{fig:stitched_image_reconstruction}
\end{figure}

\noindent $\bullet$ \textbf{\textit{How does PIS3R support downstream 3D vision tasks?}}
Our preliminary stitched image is generated via reprojection based on the camera parameters. This provides the stitched image with accurate geometric spatial information. As shown in Fig.~\ref{fig:stitched_image_reconstruction} and Sec.~\ref{sec:geometry_consistence}, stitching results of PIS3R in VLP scenarios can be directly input into a geometry-based 3D reconstruction framework for point cloud reconstruction, while better maintaining the geometric structure and integrity of the scene due to its excellent geometry preservation capability.

\noindent $\bullet$ \textbf{\textit{Limitation?}}
PIS3R, while effective in many scenarios, is not without limitations. As with most SOTA image stitching approaches, certain constraints must be acknowledged:  
\begin{itemize}
    \item \textbf{Extremely large parallax and rotation:} When both large parallax and considerable rotation are present in the input images, non-overlapping areas in either the foreground or background during the reprojection process may lead to pronounced missing regions and distortions.
\end{itemize}

\begin{itemize}
    \item \textbf{Large-area object occlusion:} When the occluded area of the background is large, the missing regions in the reprojected image lack detailed geometric structure. This effectively disables the residual diffusion process in those areas, leading to pure noise generation, which results in distortions, inconsistencies, and holes.
\end{itemize}

\section{Experiments}
In this section, we conduct a comprehensive series of comparative experiments to evaluate the performance of our method PIS3R. We compare PIS3R against four state-of-the-art image stitching solutions: \textbf{APAP}~\cite{apap_zaragoza2014projective}, \textbf{ELA}~\cite{ELA_li2018parallax}, \textbf{UDIS++}~\cite{udis2_nie2023parallax} and \textbf{OBJ-GSP}~\cite{obj-gsp_cai2025object}. 

\subsection{Datasets}
\label{sec:dataset}

PIS3R proposes a solution for image stitching under very large parallax (VLP)—a scenario not adequately addressed by existing datasets. To evaluate its effectiveness, we introduce a self-collected large-parallax dataset comprising both synthetic and real image pairs. Furthermore, to demonstrate its applicability under slight parallax, we also evaluate our method on the UDIS-D ~\cite{udis_d_nie2021unsupervised} dataset.

\subsubsection{UDIS-D Dataset} UDIS-D is a large-scale image dataset designed for image stitching or image registration. It includes image pairs with varying overlap ratios, different levels of parallax, and diverse scenarios such as indoor, outdoor, nighttime, low-light, snowy conditions, and zoom variations.
\subsubsection{Synthetic Dataset} As shown in Fig.~\ref{fig:datasets_exmaple}a, this test dataset was rendered using Blender. It comprises 20 distinct scenes \footnote{https://sketchfab.com}
\footnote{https://blenderco.cn/} \footnote{
https://free3d.com/zh/3d-models/blender} with 600 image pairs and their corresponding ground truth~(GT) images. Most image pairs have large-baseline, large-parallax characteristics.
\subsubsection{Real-World Dataset} 
\begin{figure}
    \centering
    \includegraphics[width=0.5\columnwidth]{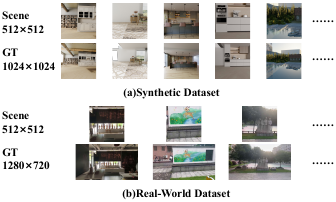}
    \caption{
    Sample images from the experimental dataset.
    (a) Synthetic dataset rendered by Blender~\cite{blender}.
    (b) Real-world dataset captured with a mobile phone.
    For each camera pose, the limited-FoV input image is center-cropped from its corresponding wide-FoV GT image and therefore shares the same camera extrinsics. Each input pair, however, consists of images acquired at two different camera poses. Consequently, the parallax is caused by relative camera translation and scene-depth variation rather than by cropping or FoV differences.}
    \label{fig:datasets_exmaple}
\end{figure}

As shown in Fig.~\ref{fig:datasets_exmaple}b, we captured the dataset with a mobile phone. It comprises 10 distinct scenes with 682 image pairs and their corresponding GT images. We test this real-world dataset to validate the performance of PIS3R in real-world scenarios.

\subsection{Comparison of Image Stitching}
\label{sec:stitching_data_experiment}

\begin{figure*}
    \centering
    \includegraphics[width=\textwidth]
    {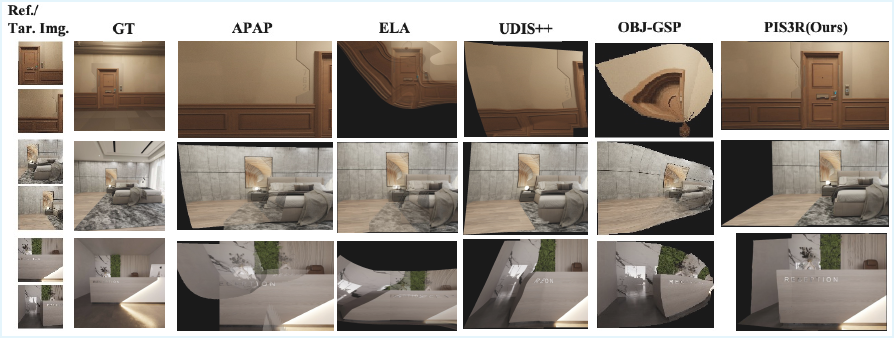}
    \caption{Qualitative comparison of stitched images on synthetic dataset.
    Our PIS3R generates images with significantly more complete structures and fewer severe distortions compared to existing image stitching algorithms APAP, ELA, OBJ-GSP and UDIS++ on the synthetic datasets. 
    Ref./Tar. Img. refers to Reference and Target Images. }
    \label{fig:blender_results}
\end{figure*}
\begin{figure*}
    \centering
    \includegraphics[width=\textwidth]
    {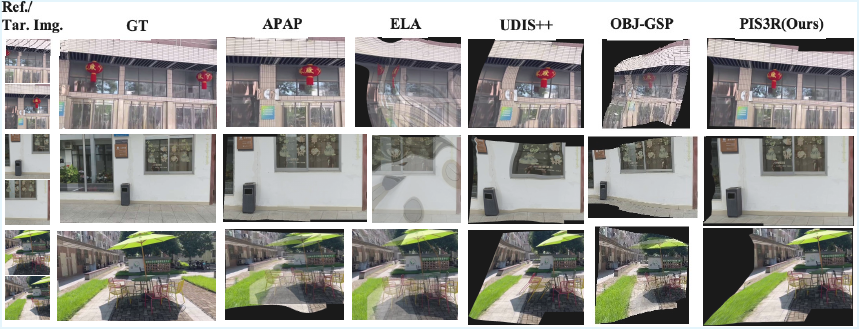}
    \caption{Qualitative comparison of stitched images on real-world dataset.
    Our PIS3R generates images with significantly more complete structures and fewer severe distortions compared to existing image stitching algorithms APAP, ELA, OBJ-GSP and UDIS++ on the real-world datasets. 
    Ref./Tar. Img. refers to Reference and Target Images. }
    \label{fig:real_results}
\end{figure*}

\begin{figure}
    \centering
    \includegraphics[width=0.7\columnwidth]
    {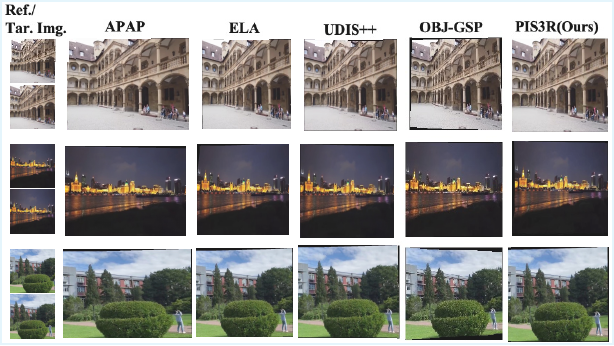}
    \caption{Qualitative comparison of stitched images on UDIS-D dataset. Even without training on the UDIS-D dataset, PIS3R remains visually comparable to various state-of-the-art methods.}
    \label{fig:UDIS-D_dataset_result}
\end{figure}

\begin{table*}
    \centering
    \begin{tabular}{ccccc}
        \toprule
        Dataset & Method & PSNR $\uparrow$ & SSIM $\uparrow$ & LPIPS $\downarrow$ \\
        \midrule
        \multirow{5}{*}{Synthetic Dataset} 
        & APAP & 19.116 & 0.659 & 0.169  \\
        & ELA & 19.813 & 0.670 & 0.158  \\
        & UDIS++ & \cellcolor[cmyk]{0,0,0,0.1} 19.858 & \cellcolor[cmyk]{0,0,0,0.1} 0.690 & \cellcolor[cmyk]{0,0,0,0.1} 0.139 \\
        & OBJ-GSP & 16.961 & 0.538 & 0.205 \\
        & PIS3R & \cellcolor[cmyk]{0,0,0,0.3} \textbf{22.026} & \cellcolor[cmyk]{0,0,0,0.3} \textbf{0.782} & \cellcolor[cmyk]{0,0,0,0.3} \textbf{0.110} \\
        \midrule
        \multirow{5}{*}{Real World Dataset} 
        & APAP & 16.764 & 0.525 & 0.246 \\
        & ELA & 16.983 & 0.547 & 0.245  \\
        & UDIS++ & \cellcolor[cmyk]{0,0,0,0.1} 17.390 & \cellcolor[cmyk]{0,0,0,0.1} 0.554 & \cellcolor[cmyk]{0,0,0,0.1} 0.204 \\
        & OBJ-GSP & 15.310 & 0.392 & 0.274 \\
        & PIS3R & \cellcolor[cmyk]{0,0,0,0.3} \textbf{19.674} & \cellcolor[cmyk]{0,0,0,0.3} \textbf{0.644} & \cellcolor[cmyk]{0,0,0,0.3} \textbf{0.180}\\
        \bottomrule
    \end{tabular}%
    \caption{Quantitative comparison of image stitching on synthetic dataset and real world dataset. PIS3R outperforms the other four methods across PSNR, SSIM and LPIPS when all image pairs are included in the evaluation.}
    \label{tab:all_results1}
\end{table*}
\begin{table}
    \centering
    \begin{tabular}{cccc}
        \toprule
        Dataset & Method & PIQE $\downarrow$ & NIQE $\downarrow$\\
        \midrule
        \multirow{5}{*}{UDIS-D}
        & APAP & 44.58& 4.383\\
        & ELA     & 47.87 & 4.152\\
         & UDIS++  & \cellcolor[cmyk]{0,0,0,0.3}\textbf{43.79} & \cellcolor[cmyk]{0,0,0,0.1}4.098 \\
        & OBJ-GSP & 56.96 & 4.885 \\
        & PIS3R   & \cellcolor[cmyk]{0,0,0,0.1}44.13 & \cellcolor[cmyk]{0,0,0,0.3}\textbf{4.059} \\
        \bottomrule
    \end{tabular}%
    \caption{Quantitative comparison of image stitching on the UDIS-D ~\cite{udis_d_nie2021unsupervised} dataset. PIS3R performs comparably to UDIS++ and both outperform other methods.}
    \label{tab:udis_d_results}
\end{table}
\subsubsection{Evaluation Metrics}We employ two sets of evaluation metrics for different datasets.

 $\bullet$ \textbf{\textit{On Dataset with GT.}} The traditional image stitching paradigm consists of warping and composition stages. However, as a 3D-reconstruction-based image stitching paradigm, PIS3R does not include these steps. Therefore, overlapping regions based metrics of warping results is not suitable for PIS3R. Notably, each reference image shares the same camera extrinsics with its GT image (Fig.~\ref{fig:datasets_exmaple}), and scene images are obtained by center-cropping GT images. Theoretically, stitched images should share the same extrinsics with reference images. Thus, we propose a novel three-step metric computation approach (Fig. D.4 in Appendix):

\textbf{Step 1:} Padding. Firstly, we pad the reference image to match the size of the GT image.

\textbf{Step 2:} Registration. Subsequently, we align the stitched image with the padded reference image using translation only, in order to compensate for differences in canvas origin and spatial offset. This step does not involve rotation, scaling, affine transformation, homography, or non-rigid deformation, and therefore does not correct geometric distortions introduced by the stitching methods.

\textbf{Step 3:} Compute metrics. We compute the evaluation metrics only over the overlapping region where both the registered stitched image and the GT image contain valid image content. Pixels belonging to padded or uncovered regions are excluded from the evaluation. The same valid-region criterion is applied to all compared methods.

We employ metrics including: \textbf{PSNR}, \textbf{SSIM} ~\cite{ssim_wang2004image}, and \textbf{LPIPS} ~\cite{lpips_zhang2018unreasonable}. 

 $\bullet$ \textbf{\textit{On Dataset without GT.}} Following the evaluation protocol of recent state-of-the-art methods OBJ-GSP, we adopt the Naturalness Image Quality Evaluator(\textbf{NIQE}) \cite{NIQE_mittal2012making} and Perception based Image Quality Evaluator(\textbf{PIQE}) \cite{PIQE_venkatanath2015blind}.\\

\subsubsection{Image Stitching in Synthetic Dataset}
\label{sec:synthetic_dataset}
 We first evaluate PIS3R and four baselines on our custom synthetic dataset.

 $\bullet$ Qualitative Comparison. The qualitative comparison results are shown in Fig.~\ref{fig:blender_results}, where the reference and target images are displayed in the left-most column. APAP, ELA, OBJ-GSP and UDIS++ exhibit noticeable artifacts and geometric inconsistencies, while Our results better preserve the geometric structure of the scene. Additional cases are provided in Fig. A.1 of Appendix.

 $\bullet$ Quantitative Comparison.
The quantitative comparison results are presented in Table~\ref{tab:all_results1}. Our PIS3R outperforms the other four methods across all metrics when including all data in the calculations. We still maintain a substantial lead in PSNR, SSIM after excluding failed registration cases as shown in Table~B.1 of Appendix. \\

\subsubsection{Image Stitching in Real-World Dataset}
\label{sec:image_stitching_in_real_scenes}
 We employ a real-world dataset captured by ourselves to validate the performance of PIS3R in real-world scenarios.

 $\bullet$ Qualitative Comparison. The qualitative comparison results are presented in Fig.~\ref{fig:real_results}. APAP, ELA, OBJ-GSP and UDIS++ exhibit noticeable artifacts and geometric inconsistencies. In contrast, our PIS3R method consistently generates images with significantly more complete structures and fewer severe distortions in these complex and challenging scenarios. Additional cases are provided in Fig. A.2 of Appendix.

 $\bullet$ Quantitative Comparison.
As shown in Table~\ref{tab:all_results1}, PIS3R demonstrates superior performance across all evaluation metrics on the complete real-world dataset. Moreover, after excluding failed registration cases, it continues to achieve leading performance on all metrics as shown in Table B.1 in Appendix.\\

\subsubsection{Image Stitching in UDIS-D Dataset}
\label{comparison on UDIS-D}
We evaluate PIS3R on the UDIS-D dataset, which primarily involves pure rotation and slight parallax.

 $\bullet$ Qualitative Comparison. The qualitative comparison results are presented in Fig. ~\ref{fig:UDIS-D_dataset_result}. Even without training on the UDIS-D dataset, PIS3R remains visually comparable to various state-of-the-art methods.

 $\bullet$ Quantitative Comparison.
As shown in Table~\ref{tab:udis_d_results}, PIS3R performs comparably to UDIS++ and outperform APAP, ELA and OBJ-GSP, even without training on the UDIS-D dataset.

\subsection{Geometric Consistency Verification of Stitched Images}
\label{sec:geometry_consistence}

In this section, we perform geometric consistency verification on stitched images of synthetic dataset to ensure structural accuracy.

$\bullet$ Evaluation Metric. Since the camera parameters of the Blender images are available, we evaluate geometric consistency by computing the Sampson epipolar error~\cite{sampson1982fitting,hartley2003multiple} ($d_{\mathrm{Sam}}^2$) between each pair of stitched images:
\begin{equation}
d_{\text{Sam}}^2 = \frac{(x_2^\top F x_1)^2}{(Fx_1)_1^2 + (Fx_1)_2^2 + (F^\top x_2)_1^2 + (F^\top x_2)_2^2},
\end{equation}
where $F$ is fundamental matrix, $x_1$ and $x_2$ respectively denote the homogeneous coordinates of a pair of corresponding feature points detected in the first and second images. Further implementation details are provided in Sec. E of the Appendix.\\
 $\bullet$ Qualitative Comparison. As shown in Fig.~\ref{fig:stitched_image_reconstruction}, we fed the stitched images of each method into AgiSoft MetaShape ~\cite{Agisoft2023} for reconstruction. PIS3R's stitching results in VLP scenarios can be directly input into a Structure From Motion(sfm) framework for point cloud reconstruction, while better maintaining the geometric structure and integrity of the scene due to its excellent geometry preservation capability.\\
$\bullet$ Quantitative Comparison.
As shown in Table~\ref{tab:epipolar_error}, PIS3R achieves a substantially lower epipolar error for stitched images exhibiting VLP conditions, notably outperforming the other four baselines.
\begin{table}
    \centering
    {
    \begin{tabular}{lcccccc}
        \toprule
        ~ & APAP & ELA & UDIS++ & OBJ-GSP & PIS3R (w/o Refine) &PIS3R(Full)\\
        $d_{\text{Sam}}^2$ $\downarrow$ & 22.78 & 38.97 & 12.94 & 16.43 &\colorbox[cmyk]{0,0,0,0.3}{\strut 2.87} &\colorbox[cmyk]{0,0,0,0.1}{\strut 4.12} \\
        \bottomrule
    \end{tabular}
    \caption{Sampson epipolar errors on the synthetic dataset.
    PIS3R (Full) denotes the final output after RDDM refinement,
    whereas PIS3R w/o Refine denotes the directly reprojected result.}
    \label{tab:epipolar_error}
    }
\end{table}

\subsection{Parallax-tolerance study}

As shown in Fig.~\ref{fig:parallax_exp}, we progressively increase the parallax by laterally translating the camera while keeping approximately the same viewing direction, with slight camera rotations allowed. In this experiment, the pure-rotation case is defined as zero parallax. Under small parallax, PIS3R and UDIS++ achieve comparable performance, whereas the advantage of PIS3R becomes increasingly evident as the parallax increases.

According to the analysis in Sec.~3.1, under pure rotation or small parallax, the inter-view correspondence can be accurately or approximately represented by 2D warping. In this regime, UDIS++ can already achieve reliable alignment. In contrast, PIS3R additionally involves camera-pose estimation, dense 3D reconstruction, and reprojection, where small estimation errors may introduce slight pixel-level deviations.

As the camera baseline increases, however, image displacement becomes increasingly dependent on scene depth, and different depth layers exhibit increasingly different motions. Such depth-dependent correspondences are difficult to represent accurately using a common 2D warping field. By explicitly reconstructing scene geometry and reprojecting points according to their 3D positions, PIS3R therefore maintains greater robustness as parallax increases.

\subsection{Ablation studies}
\label{sec:ablation_studies}

\subsubsection{Comparison of Different 3D Reconstruction Models}
We compare three feed-forward deep 3D reconstruction models: Dust3R \cite{dust3r_wang2024dust3r}, VGGT \cite{wang2025vggt}, and Mast3R \cite{Mast3r_leroy2024grounding}.

\begin{figure}
    \centering
    \includegraphics[width=0.7\columnwidth]{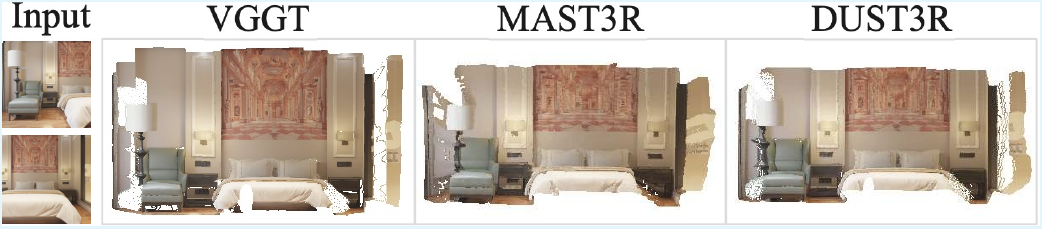}
    \caption{Qualitative comparison of different feed-forward deep 3D reoconstruction models including VGGT \cite{wang2025vggt}, MAST3R \cite{Mast3r_leroy2024grounding} and DUST3R \cite{dust3r_wang2024dust3r}. VGGT demonstrates visually superior reconstruction results.}
    \label{fig:ablation_reconstruction}
\end{figure}

$\bullet$ Qualitative Comparison. As shown in Fig.~\ref{fig:ablation_reconstruction}, VGGT demonstrates visually superior reconstruction results.

$\bullet$ Quantitative Comparison. As shown in Table~\ref{tab:ablation_reconstruction_results}, we compare the quality of reprojected images without refinement. VGGT achieves the highest reprojection results in our task

\begin{table}
    \centering
    {\small
    \begin{tabular}{ccccc}
        \toprule
        Dataset & Model & PSNR $\uparrow$ & SSIM $\uparrow$ & LPIPS $\downarrow$
         \\ \midrule
         \multirow{3}{*}{\shortstack{Synthetic \\ Dataset}}
         & Dust3R & 21.872 & 0.769 & 0.134 \\
         & Mast3R  & 21.814 & 0.780 & 0.126 \\
         & VGGT & \colorbox[cmyk]{0,0,0,0.3}{22.754} & \colorbox[cmyk]{0,0,0,0.3}{0.794} & \colorbox[cmyk]{0,0,0,0.3}{0.121} \\ 
         \midrule
         \multirow{3}{*}{\shortstack{Real World \\Dataset}}
          & Dust3R & 18.134 & 0.617 & 0.239 \\
          & Mast3R & 18.965 & 0.636 & 0.221 \\
          & VGGT & \colorbox[cmyk]{0,0,0,0.3}{20.134} & \colorbox[cmyk]{0,0,0,0.3}{0.652} & \colorbox[cmyk]{0,0,0,0.3}{0.207} \\ 
         \bottomrule
    \end{tabular}%
    }
    \caption{Ablation of feed-forward 3D reconstruction models on the synthetic and real-world datasets. Metrics are computed on preliminary reprojected images before diffusion refinement.}
    \label{tab:ablation_reconstruction_results}
\end{table}

\subsubsection{Comparison of Different Diffusion Models}
We compared three image restoration diffusion models: DiffIR ~\cite{diffir_xia2023diffir}\footnote{https://github.com/Zj-BinXia/DiffIR}, DDRM~\cite{ddrm_kawar2022denoising}\footnote{https://github.com/bahjat-kawar/ddrm}, and RDDM ~\cite{liu2024residual}\footnote{https://github.com/nachifur/RDDM}. We train DiffIR and RDDM on 3DIS-D dataset(as illustrated in Sec.~\ref{sec:Implementation}) and present comparative results of the restoration performance. 
Notably, DDRM does not rely on task-specific training and can be applied directly. 
RDDM can effectively remove noise while preserving geometric structures. 

$\bullet$ Qualitative Comparison. As shown in Fig.~\ref{fig:ablation_diffusion}, RDDM demonstrates excellent optimization results due to its unique diffusion mechanism.

$\bullet$ Quantitative Comparison. As shown in Table~\ref{tab:ablation_diffusion_results}, we quantify the metric differences between reprojected images processed and unprocessed by the diffusion model. Compared with PIS3R without refinement, RDDM introduces a modest trade-off between pixel-level fidelity and perceptual quality. On the synthetic, PSNR and SSIM decrease by 0.728 dB and 0.012, while reducing LPIPS from 0.121 to 0.110. On the real-world dataset, PSNR and SSIM decrease by 0.460 dB and 0.008, respectively, whereas LPIPS improves from 0.207 to 0.180. These results indicate that RDDM modifies local pixel values while improving perceptual similarity. As further demonstrated by the user study, the pixel-wise metrics do not fully capture the removal of reprojection noise and irregular object boundaries.

\begin{table}
    \centering
    {\small
    \begin{tabular}{ccccc}
        \toprule
        Dataset & Model & PSNR $\uparrow$ & SSIM $\uparrow$ & LPIPS $\downarrow$
         \\ \midrule
         \multirow{4}{*}{\shortstack{Synthetic \\ Dataset}}
         & w/o Refine & 22.754 & 0.794 & 0.121 \\
         & DiffIR & 21.361 & 0.740 & 0.120 \\
         & DDRM  & 20.894 &0.706 & 0.132 \\
         & RDDM & 22.026 & 0.782 & 0.110 \\ 
         \midrule
         \multirow{4}{*}{\shortstack{Real World \\Dataset}}
          & w/o Refine & 20.134 & 0.652 & 0.207 \\
          & DiffIR & 18.744 & 0.624 & 0.193 \\
          & DDRM  & 18.697 & 0.616 & 0.210 \\
          & RDDM & 19.674 & 0.644 & 0.180 \\ 
         \bottomrule
    \end{tabular}%
    }
    \caption{Ablation study of diffusion models on the synthetic and real-world datasets. RDDM improves perceptual similarity, as measured by LPIPS, while reducing PSNR and SSIM, indicating a trade-off between pixel-level fidelity and perceptual quality.}
    \label{tab:ablation_diffusion_results}
\end{table}

\begin{table}[t]
    \centering
    {\small
    \setlength{\tabcolsep}{4pt}
    \begin{tabular}{lcccc}
        \toprule
        Resolution
        & $1280\times720$
        & $720\times480$
        & $512\times512$ \\
        \midrule
        APAP\textsuperscript{1}
        & 3.46
        & 1.63
        & 1.29 \\

        ELA\textsuperscript{1}
        & 5.69
        & 2.91
        & 2.61 \\

        OBJ-GSP\textsuperscript{1}
        & 4.26
        & 2.10
        & 1.61 \\
        
        UDIS++\textsuperscript{2}
        & 0.35
        & 0.20
        & 0.18 \\

        PIS3R\textsuperscript{2}
        & 6.41
        & 3.38
        & 2.91 \\
        \bottomrule
    \end{tabular}%
    \caption{Runtime (s) comparison at three input resolutions. 1: tested with Intel i9-14900KF 3.20GHZ CPU; 2: tested with NVIDIA RTX 4090 GPU.}
    \label{tab:stitching_time}
    }
\end{table}

\subsection{Runtime and User Study}
\subsubsection{Runtime}
As shown in Table~\ref{tab:stitching_time}, since we do not have the warp and composition steps, we compare the average time required to stitch a complete image. Runtime was evaluated on 50 image pairs obtained by cropping the ground-truth images from the real-world dataset to three resolutions. Before timing at each resolution, three warm-up inference runs were performed and excluded from the measurements. Each image pair was then processed ten times. The table reports the mean runtime, excluding the one-time model-loading overhead. Since the CPU-only and GPU-based methods were tested on different hardware, the results are provided as implementation-level references rather than hardware-normalized comparisons.

\subsubsection{User Study}

\begin{figure}
    \centering
    \includegraphics[width=\textwidth]{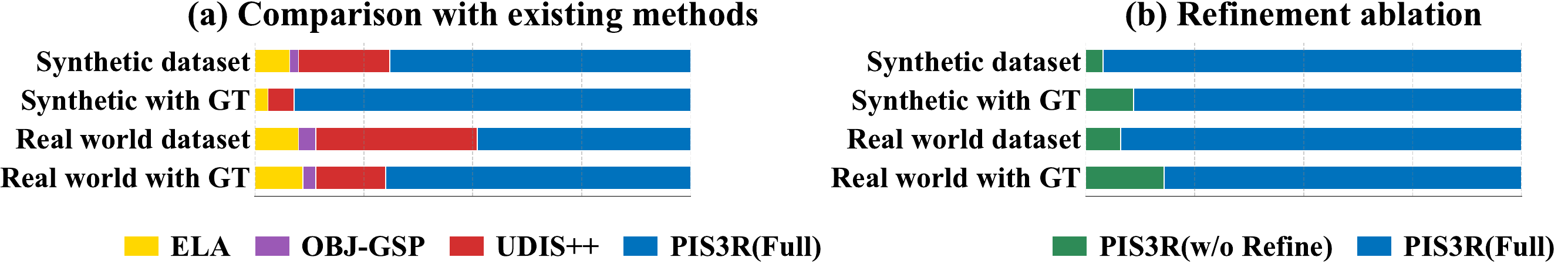}
    \caption{User-study preference rates from ten participants evaluating ten image pairs per dataset. GT was additionally shown in the ``with GT'' settings. (a) Comparison with existing methods. (b) Refinement ablation.}
    \label{fig:user_study_results}
\end{figure}

Ten participants evaluated ten image pairs per dataset, both with and without GT as an additional reference. For each pair, the input images and all compared results were shown, and participants could enlarge the images before selecting the preferred result. As shown in Fig.~\ref{fig:user_study_results}(a), PIS3R(Full) received the highest preference in every setting. In the refinement ablation, it received 82\%--96\% of the votes. Enlarged inspection showed that RDDM effectively reduced reprojection noise and irregular object boundaries. Thus, although refinement may slightly reduce PSNR and SSIM, it provides a clear perceptual improvement that is not fully captured by pixel-based metrics.
\section{Conclusion}
We proposed PIS3R, a very-large-parallax-tolerant image stitching framework based on deep 3D reconstruction. By estimating camera parameters and reconstructing a dense point cloud via a Visual Geometry Grounded Transformer, our method achieves pixel-wise alignment through reprojection. An image diffusion module further refines the reprojected result. Unlike existing methods, PIS3R preserves 3D geometric consistency, enabling seamless stitching and compatibility with downstream 3D tasks. Experimental results demonstrate that PIS3R exhibits remarkable robustness in scenes with very large parallax and outperforms prior approaches both qualitatively and quantitatively.

\section*{Acknowledgments}

This work is supported by the National Key R\&D Program of China (Grant No.~2026YFE0162500), the Project of Lushan Laboratory at Yuelushan Center for Industrial Innovation (Grant No.~2026YCII0302), Young Talents of Huxiang (Grant No.~Z202433000575), and the Changsha Science Fund for Distinguished Young Scholars (Grant No.~kq2306002).
\bibliographystyle{model1-num-names}
\bibliography{reference}

\end{document}